\documentclass[11pt]{article} 
\usepackage{rldmsubmit,palatino}
\usepackage{algpseudocode,algorithm,algorithmicx}
\usepackage{float}
\usepackage{amsmath}
\usepackage{amsfonts}
\usepackage{amssymb}
\usepackage{graphicx}
\usepackage{multicol}
\usepackage{bbm}

\algrenewcommand\algorithmicrequire{\textbf{Precondition:}}
\algrenewcommand\algorithmicensure{\textbf{Postcondition:}}
\usepackage[hidelinks]{hyperref}

\title{Efficient Reinforcement Learning via Initial Pure Exploration}

\author{
Sudeep Raja Putta \\
Conduent Labs India\\
Bangalore, India \\
\texttt{sudeepraja94@gmail.com} \\
\And
Theja Tulabandhula \\
University of Illinois at Chicago\\
Chicago,  IL 60607\\
\texttt{tt@theja.org} \\
}

\begin{document}

\maketitle

\begin{abstract}
In several realistic situations, an interactive learning agent can practice and refine its strategy before going on to be evaluated. For instance, consider a student preparing for a series of tests. She would typically take a few practice tests to know which areas she needs to improve upon. Based of the scores she obtains in these practice tests, she would formulate a strategy for maximizing her scores in the actual tests. We treat this scenario in the context of an agent exploring a fixed-horizon episodic Markov Decision Process (MDP), where the agent can practice on the MDP for some number of episodes (not necessarily known in advance) before starting to incur regret for its actions.

During practice, the agent's goal must be to maximize the probability of following an optimal policy. This is akin to  the problem of Pure Exploration (PE). We extend the PE problem of Multi Armed Bandits (MAB) to MDPs and propose a Bayesian algorithm called Posterior Sampling for Pure Exploration (PSPE), which is similar to its bandit counterpart. We show that the Bayesian simple regret converges at an optimal exponential rate when using PSPE.

When the agent starts being evaluated, its goal would be to minimize the cumulative regret incurred. This is akin to  the problem of Reinforcement Learning (RL). The agent uses the Posterior Sampling for Reinforcement Learning algorithm (PSRL) initialized with the posteriors of the practice phase. We hypothesize that this PSPE + PSRL combination is an optimal strategy for minimizing regret in RL problems with an initial practice phase. We show empirical results which prove that having a lower simple regret at the end of the practice phase results in having lower cumulative regret during evaluation.
\end{abstract}



\startmain 

\section{Introduction}

In problems involving sequential decision making under uncertainty, there exist at least two different objectives: a) optimize the online performance, and b) find an optimal behavior. In the context of Multi Armed Bandits (MAB), these objectives correspond to: a) maximize cumulative reward, and b) identify the best arm. The first objective is the widely studied problem of cumulative regret minimization. The second one is a Pure Exploration (PE) problem, where agent has to efficiently gather information to identify the optimal arm. For Markov Decision Processes (MDPs), the first objective is the classical Reinforcement Learning (RL) problem. On the other hand, PE in MDPs is an area which has not been explored in detail.

\begin{minipage}{0.75\textwidth}
Our first contribution is that we propose an algorithm called Posterior Sampling for Pure Exploration (PSPE), for PE in fixed-horizon episodic MDPs. We define an objective similar to the notion of simple regret in MABs and analyze its convergence when using PSPE. In PSPE, the agent's goal is to explore the MDP such that it maximizes the probability of following an optimal policy after some number of episodes (not necessarily known in advance). The following table captures PSPE's relation to other algorithms.
\end{minipage}\hfill
\begin{minipage}{0.24\textwidth}
\begin{center}
\begin{tabular}{| c|c|c |}
\hline
 & RL & PE \\ 
 \hline
 MAB & TS \cite{thompson1933likelihood} & PTS\cite{russo2016simple} \\  
 \hline
 MDP & PSRL \cite{osband2013more} & PSPE \\
 \hline
\end{tabular}
\label{table:1}
\end{center}
\end{minipage}

Thompson Sampling (TS) \cite{thompson1933likelihood} is a Bayesian algorithm for maximizing the cumulative rewards received in bandits. The idea is to pull an arm according to its confidence (i.e, its probability of being optimal the optimal arm). It maintains a prior distribution over bandit instances. At each step, it samples an instance of a bandit from the posterior and pulls its optimal arm. Pure exploration Thompson Sampling (PTS)\cite{russo2016simple} modifies TS by adding a re-sampling step. TS is not suitable for PE as it pulls the estimated best arm almost all the time. It takes a very long time to ascertain that none of the other arms offer better rewards. The re-sampling step prevents pulling the estimated best arm too often and helps in achieving a higher confidence in lesser number of arm pulls. Posterior Sampling for Reinforcement Learning (PSRL) \cite{osband2013more} extends TS for the RL problem on episodic fixed-horizon MDPs. It maintains a prior distribution over MDPs. At the beginning of each episode, it samples a MDP instance from the posterior, finds its optimal policy using dynamic programming and acts according to this policy for the duration of the episode. It updates the posterior with the rewards and transitions witnessed. For PE in MDPs, we propose PSPE, which adds a re-sampling step to PSRL.

In reality however, agents may have a different objective: \textbf{Optimize online performance after a period of exploration without regret}. For instance, consider a student preparing for a series of tests. She would typically take a few practice tests to know which areas she needs to improve upon. Based of the scores she obtains in these practice tests, she would formulate a strategy for maximizing her scores in the actual tests. Another example is a robot in a robotics competition. It typically has a few practice rounds before the evaluation rounds. In the practice rounds, the robot can freely explore the environment such that it maximizes its score in the evaluation round.

For this new objective, we claim that the best strategy is to use PSPE during practice and switch to PSRL during evaluation. This is our second contribution. At the end of the practice phase, PSPE maximizes the probability of following an optimal policy. It essentially initializes the priors of PSRL such that they are very close to the true MDP. PSRL can thus leverage these priors to obtain near optimal rewards.

\section{Episodic Fixed Horizon MDP}

An episodic fixed horizon MDP $M$ is defined by the tuple $\langle \mathcal{S},\mathcal{A},R,P,H,\rho \rangle$. Here $\mathcal{S}=\{ 1,...,S \} $  and  $\mathcal{A}=\{ 1,...,A \} $ are finite sets of states and actions respectively. The agent interacts with the MDP in episodes of length $H$.  The initial state distribution is given by $\rho$. In each step $h=1,...,H$ of an episode, the agent observes a state $s_h$ and performs an action $a_h$. It receives a reward $r_h$ sampled from the reward distribution $R(s_h,a_h)$ and transitions to a new state $s_{h+1}$ sampled from the transition probability distribution $P(s_h,a_h)$. The average reward received for a particular state-action is $\bar{R}(s,a) = \mathbb{E}[r|r \sim R(s,a)]$.

For fixed horizon MDPs, a policy $\pi$ is a mapping from $\mathcal{S}$ and $\{1,...,H\}$ to $\mathcal{A}$. The value of a state $s$ and action $a$ under a policy $\pi$ is:
$Q_{\pi}(s,a,h) = \mathbb{E}\bigg[\bar{R}(s,a)+ \displaystyle\sum_{i=h+1}^{H} \bar{R}(s_i,\pi(s_i,i))\bigg]$. Let $V_{\pi}(s,h) = Q_{\pi}(s,\pi(s,h),h)$. A policy $\pi^*$ is an optimal policy for the MDP if $\pi^* \in \arg \max_{\pi} V_{\pi}(s,h)$ for all $s \in \mathcal{S}$ and $h=1,...,H$. Let the set of optimal policies be $\Pi^*$.  For a MDP $M$, let $\Pi_M$ be the set of optimal policies.

\section{Posterior Sampling for Reinforcement Learning}

Consider a MDP with $S$ states, $A$ actions and horizon length $H$. PSRL maintains a prior distribution on the set of MDPs $\mathcal{M}$, i.e on the reward distribution $R$ (on $SA$ variables) and the transition probability distribution $P$ (on $S^2A$ variables). At the beginning of each episode $t$, a MDP $M_t$ is sampled from the current posterior. Let $P_t$ and $R_t$ be the transition and reward distributions of $M_t$. The set of optimal policies $\Pi_{M_t}$ for this MDP can be found using Dynamic Programming as $P_t$ and $R_t$ are known. The agent samples a policy $\pi_t$ from $\Pi_{M_t}$ and follows it for $H$ steps. The rewards and transitions witnessed during this episode are used to update the posteriors. Let $f$ be the prior density over the MDPs and $\mathcal{H}_t$ be the history of episodes seen until $t-1$. Let $s_{h,t}$ be the state observed, $a_{h,t}$ be the action performed and $r_{h,t}$ be the reward received at time $h$ in episode $t$.

\begin{minipage}[t]{.48\textwidth}

Like TS, PSRL maintains a prior distribution over the model, in this case a MDP. At each episode, it samples a model from the posterior and acts greedily according to the sample. TS selects arms according to their posterior probability of being optimal and PSRL selects policies according to the posterior probability they are optimal. It is possible to compute the posterior efficiently and sample from it by a proper choice of conjugate prior distributions or by the use of Markov Chain Monte Carlo methods. 
\begin{algorithm}[H]
  \caption{PSRL}
  \label{alg:psrl}
  \begin{algorithmic}[1]
	\State $\mathcal{H}_1= \{\}$
	\For {$t=1,2,...$}
		\State Sample $M_t \sim f(\cdot|\mathcal{H}_t)$
		\State Choose a policy $\pi_t$ at random from $\Pi_{M_t}$
		\State Observe initial state $s_{1,t}$
		\For{$h=1,...,H$}
			\State Perform action $a_{h,t} = \pi_t(s_{h,t},h)$
			\State Observe reward $r_{h,t}$ and next state $s_{h+1,t}$ 
		\EndFor
		\State $\mathcal{H}_{t+1} = \mathcal{H}_t \cup \{(s_{h,t},a_{h,t},r_{h,t},s_{h+1,t})| h=1..H\}$
	\EndFor
  \end{algorithmic}
\end{algorithm}
\end{minipage}\hfill
\begin{minipage}[t]{.5\textwidth}
\begin{algorithm}[H]
  \caption{PSPE}
  \label{alg:pspe}
  \begin{algorithmic}[1]
	\State $\mathcal{H}_1= \{\}$
	\For {$t=1,2,...$}
		\State Sample $M_t \sim f(\cdot|\mathcal{H}_t)$
		\State Sample $B \sim Bernoulli(\beta)$
		\If {$B=1$}
			\State Choose a policy $\pi_t$ at random from $\Pi_{M_t}$
		\Else
			\Repeat {}
				\State Re-sample $\widetilde{M}_t \sim f(\cdot|\mathcal{H}_t)$
			\Until $\Pi_{\widetilde{M}_t} - \Pi_{M_t} \neq \emptyset$
			\State Choose a policy $\pi_t$ at random from $\Pi_{\widetilde{M}_t} - \Pi_{M_t}$
		\EndIf
		\State Observe initial state $s_{1,t}$
		\For{$h=1,...,H$}
			\State Perform action $a_{h,t} = \pi_t(s_{h,t},h)$
			\State Observe reward $r_{h,t}$ and next state $s_{h+1,t}$ 
		\EndFor
		\State $\mathcal{H}_{t+1} = \mathcal{H}_t \cup \{(s_{h,t},a_{h,t},r_{h,t},s_{h+1,t})| h=1..H\}$
	\EndFor
  \end{algorithmic}
\end{algorithm}
\end{minipage}

\section{Posterior Sampling for Pure Exploration}

PSRL is not suitable for PE as after a certain point, it almost certainly follows the optimal policy and does not spend much effort in refining its knowledge of other policies. PSPE modifies PSRL by adding a re-sampling step. This is an extension of the Top-Two sampling idea of PTS to PSRL. This prevents it from following an estimated optimal policy too frequently.

The algorithm depends on a parameter $\beta$, where $0 < \beta < 1$, which controls how often an optimal policy of the sampled MDP is followed. At each episode $t$, PSPE samples a MDP $M_t$ and finds its set of optimal policies $\Pi_{M_t}$. With probability $\beta$ it follows a policy from this set. With probability $1-\beta$ it re-samples MDPs until a different set of policies $\Pi_{\widetilde{M}_t}$ is obtained. It then follows a policy from the set $\Pi_{\widetilde{M}_t} - \Pi_{M_t}$ for $H$ steps. In the case of bandits, PSPE is equivalent to PTS. PSRL is the same as PSPE with $\beta=1$.

\section{Analysis}

Let \(x_{\pi}(M) = {|\Pi_M \cap \{\pi\}|}/{|\Pi_M|}\). The confidence of policy $\pi$ after episode $t$ is:\ \(\alpha(\pi,t) =   \int \nolimits_{M \in \mathcal{M}} x_{\pi}(M) f(M|\mathcal{H}_{t+1}) dM\). The mean episodic reward of a policy $\pi$ is: \(\mu(\pi) = \sum_{s \in \mathcal{S}} \rho(s) V_\pi(s,1)\). 
Let $\mu^* = \max_{\pi} \mu(\pi)$.  The gap of a policy is $\Delta(\pi) = \mu^* - \mu(\pi)$ for $\pi \notin \Pi^*$. The simple regret after the episode $t$ is : \(r_t = \mu^*  -  \sum_{\pi}\alpha(\pi,t)\mu(\pi)\). Let $\Theta(t)$ be the confidence of the sub-optimal policies: \( \Theta(t) =  \sum_{\pi \notin \Pi^*}\alpha(\pi,t)\). We re-write $r_t$ as: 
\begin{align*} 
r_t &= \mu^*  - \sum_{\pi}\alpha(\pi,t)\mu(\pi)
	= \mu^* - \sum_{\pi \notin \Pi^*}\alpha(\pi,t)\mu(\pi) -  \sum_{\pi \in \Pi^*}\alpha(\pi,t)\mu^*
	= (1-\sum_{\pi \in \Pi^*}\alpha(\pi,t)) \mu^* -  \sum_{\pi \notin \Pi^*}\alpha(\pi,t)\mu(\pi)\\
	&= \Theta(t)\mu^* - \sum_{\pi \notin \Pi^*}\alpha(\pi,t)\mu(\pi)
\end{align*}
Upper and lower bounds for $r_t$ can be expressed in terms of $\Theta(t)$ and $\Delta(\pi)$:
$$
\Theta(t) \min_\pi \Delta(\pi) \leq \Theta(t)\mu^* - \sum_{\pi \notin \Pi^*}\alpha(\pi,t)\mu(\pi) \leq \Theta(t) \max_\pi \Delta(\pi)\implies \Theta(t) \min_\pi \Delta(\pi) \leq r_t \leq \Theta(t) \max_\pi \Delta(\pi)
$$
$r_t$ is bounded above and below by $\Theta(t)$ asymptotically. The convergence of $\Theta(t)$ dictates the convergence of $r_t$.

We use results from the analysis of PTS\cite{russo2016simple} about the convergence of $\Theta(t)$.

There exist constants $\{\Gamma^*_\beta > 0: \beta \in (0,1)\}$ such that $\Gamma^* = \max_\beta \Gamma^*_\beta$ exists, $\beta^* = \arg  \max_\beta \Gamma^*_\beta$ is unique for a given MDP and the following hold with probability $1$:

\begin{enumerate}
\item Under PSPE with parameter $\beta^*$,
$\displaystyle \lim_{t \to \infty} - \frac{1}{t} \log \Theta(t) = \Gamma^*$.
Under any algorithm,
$\displaystyle \limsup_{t \to \infty} - \frac{1}{t} \log \Theta(t) \leq \Gamma^*$

\item Under PSPE with parameter $\beta \in (0,1)$, $\displaystyle \lim_{t \to \infty} - \frac{1}{t} \log \Theta(t) = \Gamma^*_\beta$

\item $\Gamma^* \leq 2\Gamma^*_{\frac{1}{2}}$ and
$\displaystyle\frac{\Gamma^*}{\Gamma^*_\beta} \leq \max \bigg\{\frac{\beta^*}{\beta},\frac{1-\beta^*}{1-\beta}\bigg\}$
\end{enumerate}

$\Theta(t)$ cannot converge faster than $\exp(-t\Gamma^*)$. When using PSPE with parameter $\beta$,  $\Theta(t)$ converges at rate of $\Gamma^*_\beta$ in the limit. When $\beta = \beta^*$, this rate of convergence is $\Gamma^*$, which is optimal. When $\beta$ is close to $\beta^*$, $\Gamma^*_\beta$ is close to $\Gamma^*$. In particular, the choice of $\beta=1/2$ is robust as $\Gamma^*_{1/2}$ is atleast half of $\Gamma^*$ for any MDP.

\section{Experiments}
We compare the performance of PSPE with different values of $\beta$ and random exploration. To ease the procedure of computing posterior distributions and sampling MDPs from the posterior, we use suitable conjugate-prior distributions. For the transition probabilities, we use a uniform Dirichlet prior and a categorical likelihood, and for reward distribution, we use a Gaussian prior ($\mathcal{N}(0,1)$) and a Gaussian likelihood with unit variance. We calculate the simple regret by sampling 1000 independent MDPs from the posterior and approximating $\alpha(\pi,t)$ using sample means.  All the results are averaged across 50 trials.

Stochastic Chains [\autoref{fig:chain}], are a family of MDPs which consist of a long chain of $N$ states. At each step, the agent can choose to go left or right. The left actions (indicated by thick lines) are deterministic, but the right actions (indicated by dotted lines) result in going right with probability $1-1/N$ or going left with probability $1/N$. The only two rewards in this MDP are obtained by choosing left in state $1$ and choosing right in state $N$. These rewards are drawn from a normal distribution with unit variance. Each episode is of length $H=N$. The agent begins each episode at state $1$. The optimal policy is to go right at every step to receive an expected reward of $(1-\frac{1}{N})^{N-1}$. For the RL problem on these MDPs, dithering strategies like $\epsilon$-greedy or Boltzmann exploration are highly inefficient and could lead to regret that grows exponentially in chain length.

\begin{minipage}[t]{.49\textwidth}
\begin{figure}[H]
\includegraphics[width=\linewidth, height =\linewidth, keepaspectratio]{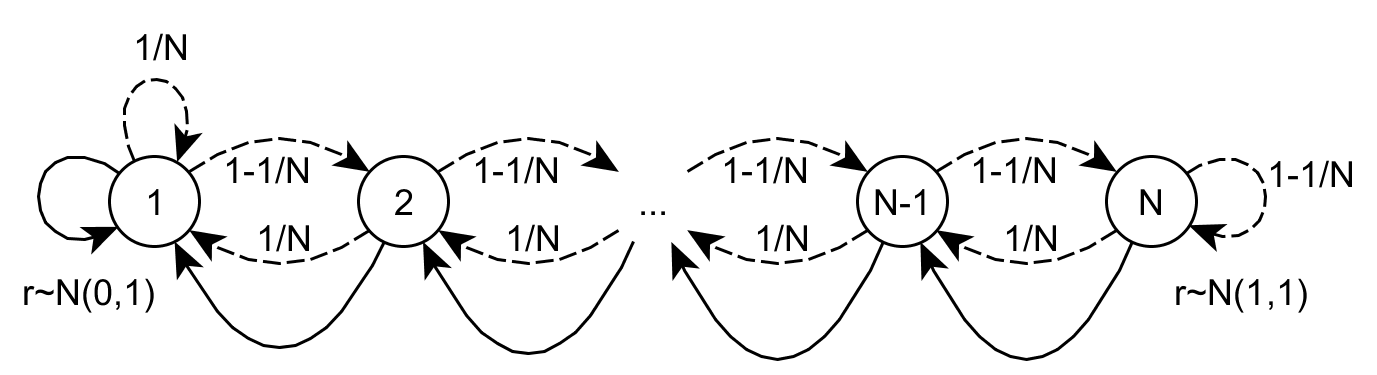}
\caption{Stochastic Chain}
\label{fig:chain}
\end{figure}
We consider a stochastic chain of length 10. The total number of deterministic policies for this MDP are $2^{10\times 10}$. We plot the simple regret of PSPE with $\beta=[0.0,0.25,0.5,0.75,0.1]$ and random exploration for $1000$ episodes in \autoref{fig:regret}. For this MDP, $\beta^*$ appears to be close to $0.25$ as the simple regret converges at the fastest rate when $\beta=0.25$. As values of $\beta$ closer to $\beta^*$ have a faster rate of convergence, the convergence of $\beta=0.0$ and $\beta=0.5$ is similar. For PSRL, which is $\beta=1.0$, the convergence is much slower. Random exploration however, is highly inefficient. This is because PSPE is able to achieve ``Deep Exploration" \cite{osband2016deep} whereas random exploration does not. Deep Exploration means that the algorithm selects actions which are oriented towards positioning the agent to gain useful information further down in the episode.
\end{minipage}\hfill
\begin{minipage}[t]{.49\textwidth}
\begin{figure}[H]
\includegraphics[width=\linewidth, height =\linewidth,keepaspectratio]{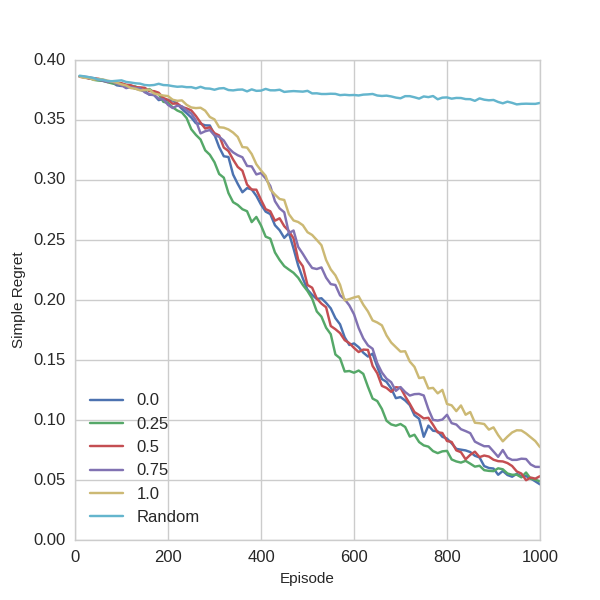}
\caption{Simple Regret of Stochastic Chain}
\label{fig:regret}
\end{figure}
\end{minipage}

\section{Reinforcement Learning with Practice}
In this section, we try to answer the question: ``When does it make sense for the agent to use PSPE?".
Consider the following situation: The agent's goal is to maximize the cumulative reward, but the rewards are accumulated from the $T$th episode. The rewards obtained during episodes $1$ to $T-1$ are not used to evaluate the agent's performance. 

\begin{minipage}{.49\textwidth} 
The first $T-1$ episodes can be considered as practice, where the agent gathers information so that it obtains near optimal rewards from episode $T$. The agent may not know $T$ in advance. It will be told at the beginning of the $T$th episode that its performance is being evaluated. 
It is not entirely apparent which strategy the agent should use during practice. The agent could ignore the fact that the rewards accumulated during practice do not matter and always use a reward maximizing strategy such as PSRL. We argue that the best strategy is to use PSPE during practice and switching to PSRL during evaluation. Logically, having lower simple regret after practice should result in lower cumulative regret during  evaluation. Since PSPE with parameter $\beta^*$ reaches a lower simple regret faster than PSRL, an optimal policy of a sampled MDP will be an optimal policy of the true MDP with high probability. Hence, we claim that lower regret is incurred by PSRL in the evaluation when PSPE is used during practice.

Like before, we consider a stochastic chain of length 10. We let the practice phase last for different intervals starting from $0$ to $1000$ in steps of 10. During practice, the agents use PSPE with with $\beta=[0.0,0.25,0.5,0.75,0.1]$. After practice, the agents use PSRL for 1000 episodes. The cumulative regret of these agents after the 1000 episodes is plotted against the simple regret at the end of practice. The simple and cumulative regrets are highly correlated, as show in \autoref{fig:simple_regret}.
\end{minipage}\hfill
\begin{minipage}{.49\textwidth}
\centering
\begin{figure}[H]
\includegraphics[width=\linewidth, height =\linewidth,keepaspectratio]{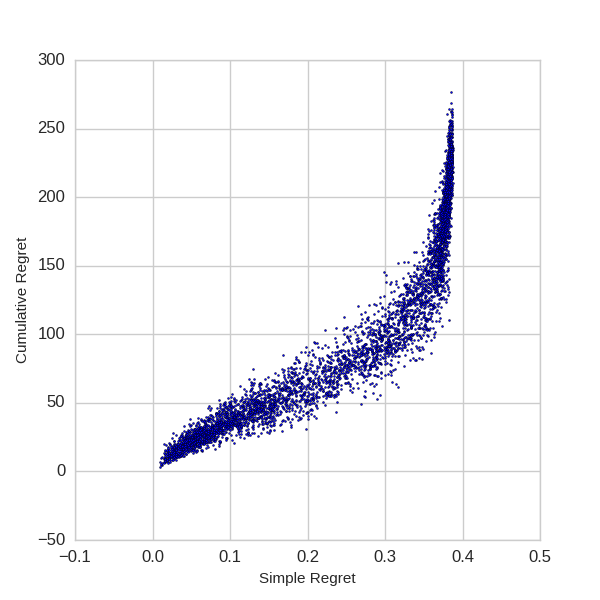}
\caption{Simple Regret vs Cumulative Regret}
\label{fig:simple_regret}
\end{figure}
\end{minipage}

\section{Conclusion and Future Work}
In this paper, we present PSPE, a Bayesian algorithm for the Pure exploration problem in episodic fixed-horizon MDPs. PSPE combines the Top-Two sampling procedure of PTS with PSRL. We define a notion of simple regret and show that it converges at an optimal exponential rate when using PSPE. Using stochastic chain MDPs, we compare the convergence of simple regret for PSPE with various values of parameter $\beta$. We also define the practical problem of Reinforcement Learning with practice. We empirically show that a combination of PSPE and PSRL can offer a feasible solution for this problem. We intend to further explore the problem of RL with practice and provide theoretical guarantees in the case of bandits and MDPs.

PSPE requires solving MDPs through dynamic programming at each step. An alternative approach, which avoids solving sampled MDPs is value function sampling \cite{dearden1998bayesian}. Using value function sampling approaches to achieve pure exploration remains an open research direction.

\bibliographystyle{named}

\end{document}